\documentclass{article}
\usepackage{spconf,amsmath,graphicx, kotex}
\usepackage{verbatim}
\usepackage{graphicx}
\usepackage{subcaption}
\usepackage{booktabs}
\usepackage{cite}
\usepackage{ mathrsfs }

\title{AD-VO: Scale-resilient visual odometry using attentive disparity map}
\name{Joosung Lee, Sangwon Hwang, Kyungjae Lee, Woo Jin Kim, Junhyeop Lee, Tae-young Chung, Sangyoun Lee\thanks{This work was supported by Institute for Information and communications Technology Promotion(IITP) grant funded by the Korea government (MSIP)(No.2016-0-00197, Development of the high-precision natural 3D view generation technology using smart-car multi sensors and deep learning)}}
\address{School of Electrical and Electronic Engineering, Yonsei University, Seoul, Republic of Korea\\E-mail:\{m3155, sangwon1042, kjaelee, woojinkim0207, jun.lee, tato0220, syleee\}@yonsei.ac.kr}

\begin{document}
	%
	\maketitle
	\begin{abstract}
		Visual odometry is an essential key for a localization module in SLAM systems. However, previous methods require tuning the system to adapt environment changes. In this paper, we propose a learning-based approach for frame-to-frame monocular visual odometry estimation. The proposed network is only learned by disparity maps for not only covering the environment changes but also solving the scale problem. Furthermore, attention block and skip-ordering scheme are introduced to achieve robust performance in various driving environment. Our network is compared with the conventional methods which use common domain such as color or optical flow. Experimental results confirm that the proposed  network shows better performance than other approaches with higher and more stable results.
	\end{abstract}
	\section{Introduction}
	\label{sec:intro}
	
	In Simultaneous Localization and Mapping (SLAM) system, it is important to acquire the position and orientation of the camera called visual odometry (VO). Generally, VO in SLAM systems has been studied through feature-based method~\cite{mur2015orb,klein2007parallel,davison2007monoslam} and direct-based method~\cite{engel2014lsd}. These methods focus on camera trajectory optimization while not on frame-to-frame (F2F) estimation.
	
	Our approach aims to estimate F2F VO without optimization methods, such as loop closing~\cite{mur2015orb} and bundle adjustment~\cite{engels2006bundle}. Geiger et al.~\cite{geiger2011stereoscan} suggested the feature-based F2F VO which generates consistent 3D point cloud through feature matching between frames. Ciarfuglia et al.~\cite{ciarfuglia2014evaluation} presented the correlation between optical flow and camera ego-motion in a non-geometric method by adopting support vector machine.
	\begin{figure}[!t]	
		\centering
		\centerline{\includegraphics[width=7.5cm]{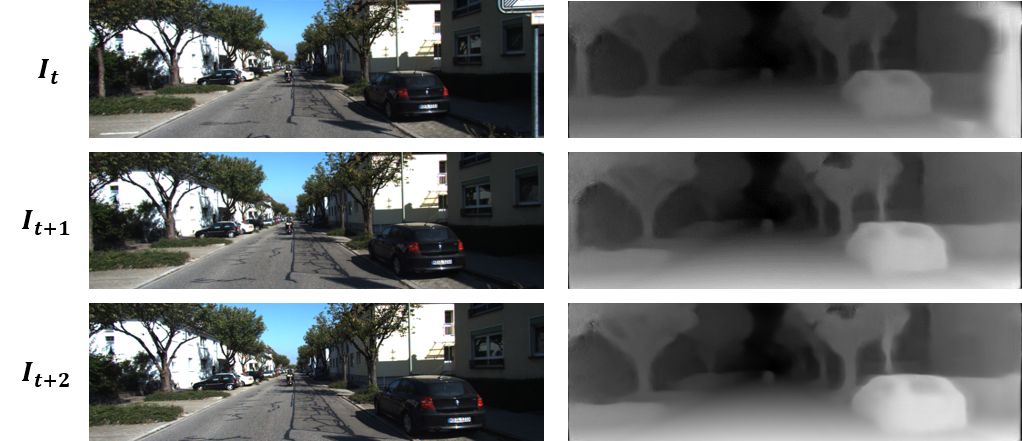}}
		\caption{RGB images~\cite{geiger2012we} and disparity maps~\cite{godard2017unsupervised} (images are on temporal order from top to bottom).}
		\label{fig:rgb2dis}
	\end{figure}

	Recently, deep networks showed remarkable improvement of the computer vision technology~\cite{redmon2016you, lee2017brightness}. One of the first convolutional neural network (CNN) based methods were proposed by Konda et al.~\cite{konda2015learning, konda2013unsupervised}, who showed the feasibility of learning F2F visual odometry. Moreover, Costante~et~al.~\cite{costante2016exploring} (P-CNN) and Muller~et~al.~\cite{muller2017flowdometry} (Flowdometry) predicted F2F camera ego-motion using pre-built optical flow images~\cite{brox2004high, dosovitskiy2015flownet}. They insisted that using optical flow images as training is adequate than RGB domain because it contains displacement information. However, these methods still remain a scale problem. SFMLearner~\cite{zhou2017unsupervised} and UndeepVO~\cite{li2017undeepvo} predict depth information and camera ego-motion through unsupervised learning. Considering the potentials of deep learning-based methods, we design an end-to-end deep convolution neural network to estimate F2F monocular visual odometry.

	The contributions of our paper are summarized as follows. First, our visual odometry network is implemented using only the disparity map (right column of Fig.~\ref{fig:rgb2dis}). Since the disparity map has spatial clues in each frame, we can effectively address the scale problem and obtain better performance than the current existing optical flow-based networks such as P-CNN~\cite{costante2016exploring} and Flowdometry~\cite{muller2017flowdometry}. In addition, our network was designed to extract both attention and feature map through two parallel blocks. The attention block enables to focus on sensitive regions of the image. Second, a skip-ordering scheme is introduced to learn larger displacements of frames by training additional image pairs. The contributions of our method do not only improve the performance but also enable the camera ego-motion to have robustness regardless of diverse driving environment.	
	
	\begin{figure*}[!t]
		\centering
		\includegraphics[scale=0.333]{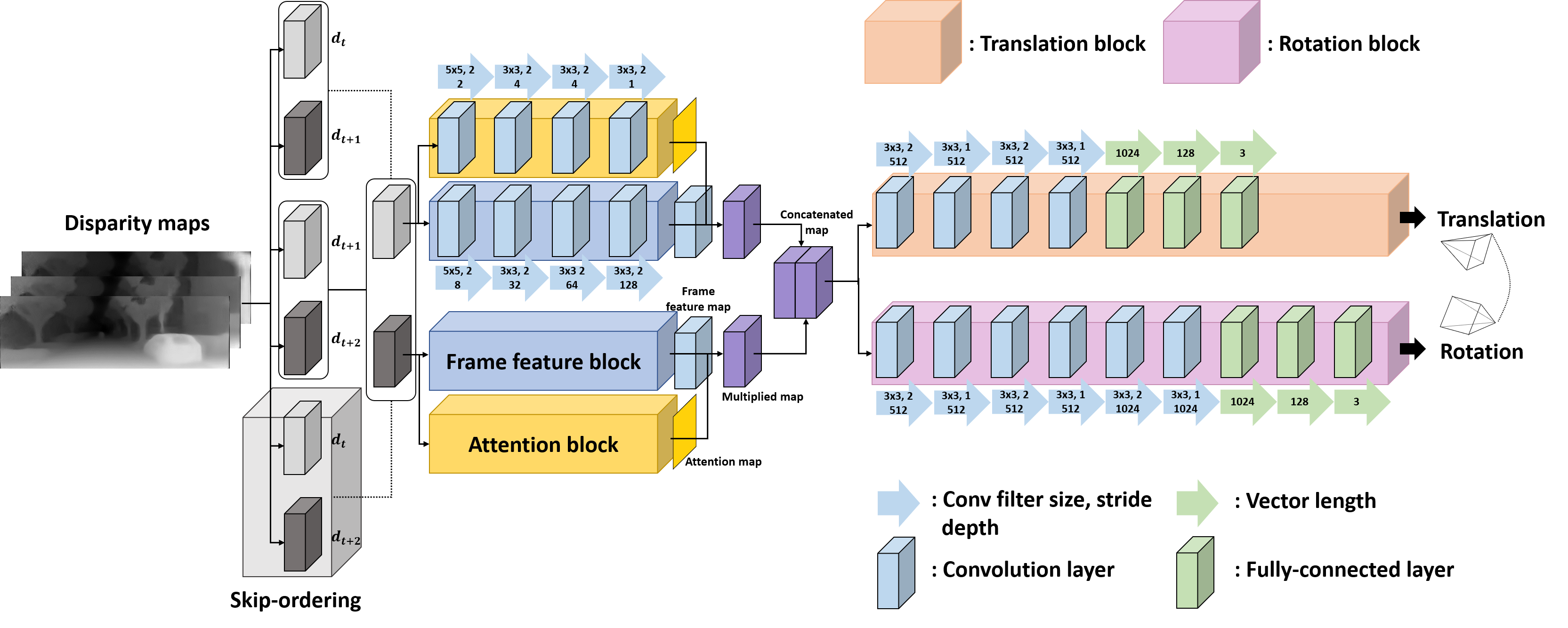}
		\caption{Frame to frame odometry model which is based on disparity maps. Our network has four blocks of `Frame feature', `Attention', `Translation' and `Rotation'.}
		\label{fig:model}
	\end{figure*}
	
	\section{Proposed method}
	\label{sec:proposed method}
	In this section, we describe a deep learning network that predicts F2F camera ego-motion, as illustrated in Fig.~\ref{fig:model}. The proposed network focuses on improving odometry estimation accuracy based on disparity maps with skip-ordering (SO) strategy. In this study, monocular depth estimation based on~\cite{godard2017unsupervised} is employed to generate disparity maps. The network consists of four blocks: frame feature, attention, translation, and rotation. The details of the proposed network are explained in the following subsections.
	
	\subsection{Network architecture}
	\label{sec:Network architecture}
	The front part of the network is designed to extract frame feature and attention maps through two parallel blocks, as illustrated in Fig.~\ref{fig:model}. A frame feature map contains general information about a camera ego-motion prediction. An attention map reflects the sensitivity of the camera ego-motion estimation in the frame feature map. Since intensity value in the disparity map is highly dependent on the camera motion displacement, the attention map is learned by utilizing its property. For instance, camera ego-motion is difficult to be predicted from objects in a long distance such as buildings, trees, and sky in the scene. After each extraction, the attention map and frame feature map are pixel-wisely multiplied per frame, and then two multiplied maps are concatenated.
	
	Translation and rotation blocks use the concatenated map as their input. Each block learns rotation and translation information in parallel to be robust on feature learning. The rotation block has two additional layers than the translation block because the former has higher nonlinearity than the latter~\cite{kendall2015posenet}. Moreover, we normalize losses to balance rotation and translation errors, which is described in detail in the next section.
	
	All layers use the ReLU as an activation function except for the attention layer. Each attention layer is followed by the sigmoid activation function to yield the value between 0 and 1 for the attention map.
		
	\subsection{Training and testing}
	\label{sec:train and test}
	The KITTI dataset~\cite{geiger2012we} is used in train~(sequence number 00 to 07) and test~(08 to 10) phases~\cite{costante2016exploring,muller2017flowdometry}.
	The ground truth of the KITTI has 12 values per an image, which of 9 are the rotation matrix and 3 are the position. These are the information about the first frame of the sequences in world coordinate system. Therefore, it is necessary to change the related information between frames, to train the network. Eq.~\ref{eq:eq1} below explains how the information about rotation matrix \textit{R}, position \textit{P} and translation \textit{T} is obtained, which are expressed as:
	\begin{align}
	\begin{split}
	R_{ji}&=R_{i1}^{-1}\times R_{j1},\\
	P_{ji}&=P_{j1}-P_{i1},\\
	T_{ji}&=R_{i1}^{-1}\times P_{ji},
	\end{split}
	\label{eq:eq1}
	\end{align}
	where $R\in \mathbf{SO(3)}$ and $P, T\in \mathbf{R^{3}}$ of the $j$-th frame with respect to the $i$-th frame.
	
	A rotation matrix is generally represented by Euler angle, quaternion or Lie group (3). However, we empirically found that the quaternion vectors have the limitations in learning the rotation matrix. 
	When the amount of rotation is small, one of four elements in the quaternion has abnormally higher value than the others, which can be subjected as a biased result. In Lie group (3), one-to-one mapping between rotation matrix and Lie group (3) cannot be transformed with each other when no rotation occurs. Therefore, Euler angle is suitable for F2F camera ego-motion estimation in our network.
	
	\subsubsection{Training}
	\label{sec:Training}
	
	In training phase, three consecutive disparity maps are used as an input ($d_{t}$, $d_{t+1}$, $d_{t+2}$), where \textit{d} represents the disparity map, and \textit{t} denotes the time order. 
	As described above, skip-ordering scheme is introduced to learn larger displacements of frames by training additional image pairs. Thus, we use ($d_{t}$, $d_{t+2}$) as skip-ordering~(SO) and each elements in ($d_{t}$, $d_{t+1}$), ($d_{t+1}$, $d_{t+2}$), ($d_{t}$, $d_{t+2}$) are paired with others. Through SO strategy, the network becomes robust against various motion changes.
	
	Our network loss consists of the weighted sums of rotation and translation parts, which can be expressed as: 
	\begin{align}
	\begin{split}
	L^{R}&=L_{21}^{R}+L_{32}^{R}+L_{31}^{R},\\
	L^{T}&=L_{21}^{T}+L_{32}^{T}+L_{31}^{T},\\
	\mathscr{L}&=\alpha  L^{R}+L^{T},
	\end{split}
	\label{eq:loss_total}
	\end{align}	
	where $L^{R}$ and $L^{T}$ are rotation and translation losses, respectively. The weighted factor $\alpha$ is known to have a large value to normalize the losses of rotation and translation, because the former has higher nonlinearity than the latter~\cite{kendall2015posenet}. We experimentally set $\alpha=350$. The mean squared error \textit{L} is described in Eq.~\ref{eq:loss_L2}. 
	
	\begin{equation}
	\begin{split}
	L=\frac{1}{N} \sum_{i=1}^{N}\left \| \textbf{GT} - \textbf{output} \right \|_{i}^2,
	\end{split}
	\label{eq:loss_L2}
	\end{equation}
	in which \textit{N} is the number of training samples.
	
	The loss~$\mathscr{L}$ is minimized by the Adam optimization~\cite{kingma2014adam}. The learning rate starts from 1$e-$5, then reduces by a factor of 2 for every 5 epochs until 30 epochs.
	
	\subsubsection{Testing}
	\label{sec:Testing}
	The network uses two consecutive disparity maps as an input and yields the relation of translation and Euler angle between the frames.

	To obtain the rotation matrix and position vector from $1$st to $j$-th frame, following equations are required:
	 
	\begin{align}
	\begin{split}
	P_{j1}&=P_{i1}+R_{i1}\times T_{ji},\\
	R_{j1}&=R_{i1}\times R_{ji},
	\end{split}
 	\label{eq:eq3}
	\end{align}

	The rotation matrix of the first frame~($R_{11}$) is set to 3x3 identity matrix and located at the origin of world coordinates, (0, 0, 0). The obtained results are evaluated using the provided code by KITTI~\cite{geiger2012we}.
	
	\section{Experiments}
	\label{sec:pagestyle}   	
	The proposed method has been compared with the handcraft-based method VISO2-M~\cite{geiger2011stereoscan}, and three learning-based methods of SVR-VO~\cite{ciarfuglia2014evaluation}, P-CNN~\cite{costante2016exploring}, and Flowdometry~\cite{muller2017flowdometry}.
	
	All methods were evaluated using KITTI benchmark metrics. The VISO2-M and SVR-VO were implemented by the provided code in~\cite{geiger2011stereoscan,costante2016exploring}, receptively. Each result of P-CNN and Flowdometry is reported in ~\cite{costante2016exploring,muller2017flowdometry}. Since VISO2-M does not solve the scale problem, we recovered the scale through the range of position in ground truth. Table.~\ref{tab:results_others} shows the performance of the compared algorithms. 
				
	To evaluate the structure of our network, RGB-VO and D-VO have been additionally experimented. Each network was trained end-to-end by monocular RGB and disparity map using SO without attention layers. As presented in Table.~\ref{tab:results_ours}, RGB-VO shows higher average translation performance than compared to VISO2-M and SVR-VO. However, it is worse than other deep network approaches. D-VO, which simply replaced the domain with a disparity map, has better average translation performance than RGB-VO. Translation error is reduced from 13.83\% to 12.44\% indicating that using disparity map is effective on translation accuracy. However, since a significant improvement of rotation is not found, merely using disparity map is not adequate. 
	
	For evaluating the value of an attention block, AD-VO with SO was additionally evaluated. From the comparison of D-VO and AD-VO with SO, translation error was reduced from 12.44\% to 8.59\% and the rotation error was reduced from 0.0474\% to 0.0334\%. These results prove that the usage of the attention block is effective. Accordingly, AD-VO with SO shows the best average error on translation and performance of rotation accuracy in sequence 10. The further experiment on D-VO was conducted without SO to determine the effect of SO. Comparing D-VO without and with SO, the translation error was reduced from 16.02\% to 12.44\%, with the rotation error from 0.0562\% to 0.0474\%. From the comparison of the result, it can be seen that using attention layer and SO improve the performance.    
	
\begin{figure*}[!t]
		\begin{subfigure}[b]{0.32\linewidth}
			\centering
			\includegraphics[width=0.26\textheight]{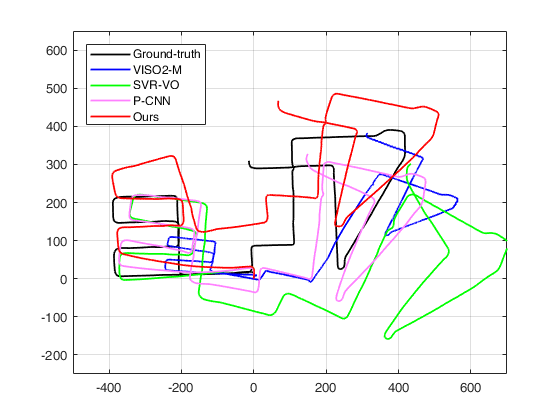}
			\caption{Sequence 08}
		\end{subfigure}
		\begin{subfigure}[b]{0.32\linewidth}
			\centering
			\includegraphics[width=0.26\textheight]{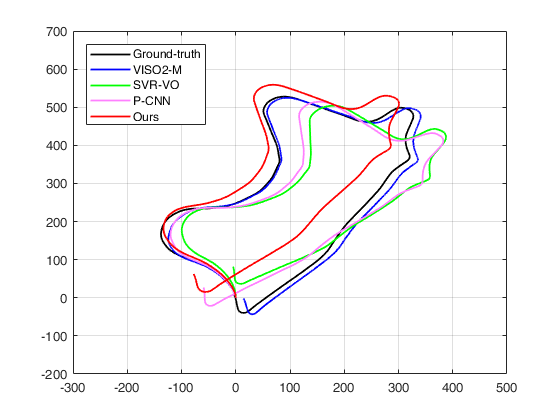}
			\caption{Sequence 09}
		\end{subfigure}
		\begin{subfigure}[b]{0.32\linewidth}
			\centering
			\includegraphics[width=0.26\textheight]{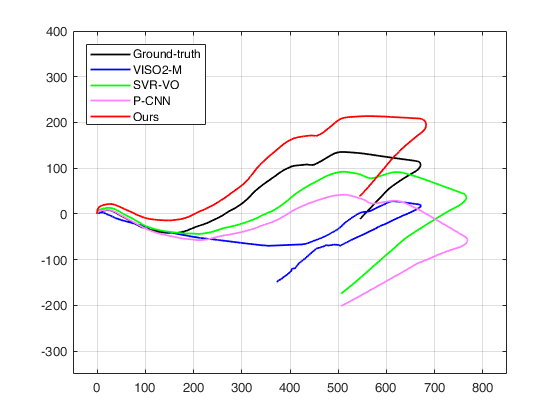}
			\caption{Sequence 10}
		\end{subfigure}
		\caption{Reconstructed trajectories of test sequences(08, 09, 10). Our algorithm is  AD-VO with skip-ordering.}
		\label{fig:odometry}
	\end{figure*}

	Attention layer and SO does not only improve the performance but also stabilize the result. We have analyzed the standard deviation of each algorithm and network between the test sequences. The standard deviation of AD-VO with SO was measured as the smallest among the algorithms with 0.868 and 0.001 in translation and rotation. Unlike the other networks, our method have small variation in translation error between test sequences and stable results, regardless of the environment. Through an analysis of the attention block, it determines which regions should be considered. Fig.~\ref{fig:odometry} shows the result of reconstructed trajectories in test sequences.
	
	\begin{table*}[!t]
		\centering
		\resizebox{\textwidth}{!}{
			\begin{tabular}{|c||c|c|c|c|c|c|c|c|}
				\hline
				& \multicolumn{ 2}{|c|}{VISO2-M} & \multicolumn{ 2}{|c|}{SVR-VO} & \multicolumn{ 2}{|c|}{Flowdometry} & \multicolumn{ 2}{|c|}{P-CNN VO} \\
				\hline
				Seq & Trans [\%] & Rot [deg/m] & Trans [\%] & Rot [deg/m] & Trans [\%] & Rot [deg/m] & Trans [\%] & Rot [deg/m] \\
				\hline
				08 &     26.213 &     0.0247 &      15.42 &     0.0363 &       9.98 &     0.0544 &       7.60 &     0.0187 \\
				\hline
				09 &       4.09 &     0.0124 &      10.50 &     0.0445 &      12.64 &     0.0804 &       6.75 &     0.0252 \\
				\hline
				10 &      60.02 &     0.0669 &      21.97 &     0.0545 &      11.65 &     0.0728 &      21.23 &     0.0405 \\
				\hline
				\hline
				avg &      24.91 &     0.0266 &      15.02 &     0.0401 &      10.77 &     0.0623 &       8.96 &     0.0235 \\
				\hline
				std &     15.305 &      0.015 &      3.165 &     0.0061 &      1.136 &     0.0113 &      4.689 &     0.0067 \\
				\hline
			\end{tabular}  		                      
		}
		\caption{The performance of the comparison algorithms. Translation and rotation error of test sequences using KITTI devkit.}
		\label{tab:results_others}
	\end{table*}

	\begin{table*}[!t]
		\centering
		\resizebox{\textwidth}{!}{
			\begin{tabular}{|c||c|c|c|c|c|c|c|c|}
				\hline
				& \multicolumn{ 2}{|c|}{RGB-VO} & \multicolumn{ 4}{|c}{D-VO} & \multicolumn{ 2}{|c|}{AD-VO} \\
				\hline
				& \multicolumn{ 2}{|c|}{w/ skip-ordering} & \multicolumn{ 2}{|c|}{w/o skip-ordering} & \multicolumn{ 2}{|c|}{w/ skip-ordering} & \multicolumn{ 2}{|c|}{w/ skip-ordering} \\
				\hline
				Seq & Trans [\%] & Rot [deg/m] & Trans [\%] & Rot [deg/m] & Trans [\%] & Rot [deg/m] & Trans [\%] & Rot [deg/m] \\
				\hline
				08 &      14.39 &     0.0452 &      16.41 &     0.0547 &      13.36 &     0.0478 &       9.21 &     0.0341 \\
				\hline
				09 &       9.59 &     0.0397 &      13.70 &     0.0417 &       8.79 &     0.0365 &       7.43 &     0.0320 \\
				\hline
				10 &      19.18 &     0.0438 &      18.48 &     0.0954 &      14.36 &     0.0671 &       7.26 &     0.0317 \\
				\hline
				\hline
				avg &      13.83 &     0.0438 &      16.02 &     0.0562 &      12.44 &     0.0474 & {\bf 8.59} & {\bf 0.0334} \\
				\hline
				std &      2.718 &     0.0023 &      1.407 &     0.0147 &      1.995 &     0.0083 & {\bf 0.868} & {\bf 0.001} \\
				\hline
			\end{tabular}  			
			  					                                             
		}
		\caption{The performance of our algorithms. Translation and rotation error of test sequences using KITTI devkit. AD-VO works reliably in a variety of driving environment. `D' means that we use disparity map and `A' means that model adapt attention block.}
		\label{tab:results_ours}
		
	\end{table*}
	
	\section{Conclusion}
	\label{sec:typestyle}
	In this paper, we presented a novel system to obtain F2F camera ego-motion from monocular images. We studied four different algorithms and compared the performance using evaluation metrics provided by KITTI benchmark. Our system is designed not only to recover the scale problem but also to achieve stable results in various environment. An Attention block allows where the network should be trained to focus on influential regions of the image. Moreover, we suggested a skip-ordering scheme to train the larger displacement of camera ego-motion. To the best of our knowledge, DA-VO is the first F2F camera ego-motion network using only disparity maps. In the future, we aim to enhance the rotation performance by combining optical flow and disparity map. 
	
	\vfill
	\pagebreak
	
	\label{sec:ref}
	
	\bibliographystyle{IEEEbib}
	\bibliography{refs}
	
\end{document}